\documentclass[11pt]{article}

\usepackage{microtype} 
\usepackage{booktabs}  
\usepackage{url}  

\usepackage{amsmath}
\usepackage{amsthm}

\usepackage[preprint]{automl}
\usepackage{threeparttable}
\usepackage{booktabs}
\usepackage{tabularx}

\usepackage[nolist]{acronym}
\begin{acronym}
    \acro{QAT}{Quantization-Aware Training}
    \acro{PTQ}{Post-Training Quantization}
    \acro{AIMC}{Analog In-Memory Computing}
    \acro{NAS}{Neural Architecture Search}
    \acro{DAG}{Directed Acyclic Graph}
    \acro{SGD}{Stochastic Gradient Descent}
    \acro{HWT}{Hardware-Aware Training}
    \acro{NN}{Neural Network}
    \acro{DNN}{Deep Neural Network}
\end{acronym}

\usepackage{natbib}
\title{AnalogNAS-Bench: A NAS Benchmark for Analog In-Memory Computing}

\author[1,$\ast$]{\nameemail{Aniss Bessalah}{ka_bessalah@esi.dz}}
\author[1,$\ast$]{\nameemail{Hatem Mohamed Abdelmoumen}{kh_abdelmoumen@esi.dz}}
\author[1, 2]{\nameemail{Karima Benatchba}{k_benatchba@esi.dz}}
\author[3]{\nameemail{Hadjer Benmeziane}{hadjer.benmeziane@ibm.com}}

\affil[$\ast$]{Equal contribution.}
\affil[1]{Ecole Nationale Supérieure d’Informatique, 16309 Oued Smar, Algiers, Algeria}
\affil[2]{Laboratoire de Méthodes de Conception des Systèmes, 16309 Oued Smar, Algiers, Algeria}
\affil[3]{IBM Research Europe, 8803 Rüschlikon, Switzerland}

\hypersetup{%
  pdfauthor={}, 
  pdftitle={},
  pdfsubject={},
  pdfkeywords={}
}

\begin{document}

\maketitle

\begin{abstract}
Analog In-memory Computing (AIMC) has emerged as a highly efficient paradigm for accelerating \acp{DNN}, offering significant energy and latency benefits over conventional digital hardware. 
However, state-of-the-art neural networks are not inherently designed for AIMC, as they fail to account for its unique non-idealities. Neural Architecture Search (NAS) is thus needed to systematically discover neural architectures optimized explicitly for AIMC constraints. However, comparing NAS methodologies and extracting insights about robust architectures for AIMC requires a dedicated NAS benchmark that explicitly accounts for AIMC-specific hardware non-idealities.  
To address this, we introduce AnalogNAS-Bench, the first NAS benchmark tailored specifically for AIMC. Our study reveals three key insights: (1) standard quantization techniques fail to capture AIMC-specific noises, (2) robust architectures tend to feature wider and branched blocks, (3) skip connections improve resilience to temporal drift noise. These insights highlight the limitations of current NAS benchmarks for AIMC and pave the way for future analog-aware NAS.
All the implementations used in this paper can be found at \url{https://github.com/IBM/analog-nas/tree/main/analognasbench}.
\end{abstract}


\section{Introduction}

\acp{DNN} deployment is increasingly constrained by power consumption and memory bandwidth limitations. \ac{AIMC} \citep{sebastian2020memory} has emerged as a promising alternative to traditional digital hardware by performing computations directly within memory arrays, reducing data movement overhead and significantly improving energy efficiency and computational throughput. By leveraging memory devices, \ac{AIMC} enables matrix-vector multiplications to be executed in a single step \citep{lammie2022memtorch}, offering orders-of-magnitude improvements in energy efficiency compared to conventional Von Neumann platforms. 

Despite its advantages, \ac{AIMC} introduces several non-idealities that impact \ac{DNN} performance \citep{boybat2021temperature}. Unlike digital accelerators, which rely on precise arithmetic operations, \ac{AIMC} suffers from device-to-device variations, cycle-to-cycle noise, temporal drift, and limited precision due to analog-digital conversion. These imperfections lead to unpredictable accuracy degradation when deploying conventional \acp{DNN} on \ac{AIMC} platforms. As a result, models that perform well in digital settings often fail to retain their accuracy when executed on \ac{AIMC}. 

To address this challenge, researchers have explored two main directions. One approach to mitigate \ac{AIMC} non-idealities' effect is to modify the training process by injecting hardware-specific noise and variations into the learning pipeline. This technique, commonly known as \textbf{\ac{HWT}} \citep{rasch2023hardware}, allows the model to adapt to \ac{AIMC}-induced errors by learning robust representations. An alternative strategy is to design \acp{DNN} that are inherently robust to \ac{AIMC} noises. \textbf{\ac{NAS}} \citep{elsken2019neural} automates the discovery of efficient network topologies by searching for architectures that maximize accuracy while considering hardware constraints. 

However, despite the progress in \ac{HWT} and the numerous analog-aware \ac{NAS} methodologies, fundamental questions remain: (1) What architectural characteristics enable certain networks to recover their digital accuracy after \ac{HWT}, while others do not? (2) Can NAS methodologies developed for quantization also be leveraged for \ac{AIMC}, thereby avoiding the need for costly \ac{HWT}? (3) How to systematically compare among \ac{NAS} methodologies for \ac{AIMC}?

\noindent Answering these questions requires a systematic evaluation of different architectures under AIMC-specific constraints. Existing NAS benchmarks, such as NAS-Bench-101 \citep{ying2019bench} and NAS-Bench-201 \citep{dong2020nasbench201}, typically evaluate architectures based on digital accuracy, FLOP count, and latency, but they do not incorporate \ac{HWT}, which is crucial for assessing model robustness under \ac{AIMC} conditions. As a result, researchers lack a standardized benchmark to analyze which architectures best withstand AIMC-induced errors and compare AIMC-targeted NAS methodologies.

To bridge this gap, we introduce \textbf{AnalogNAS-Bench}, the first NAS benchmark specifically designed for AIMC. Our benchmark enables a structured evaluation of architectures under AIMC constraints, using AIHWKit \citep{aihwkit}, and provides insights into architectural robustness when combined with \ac{HWT}.  Our key contributions are:
\begin{enumerate}
    \item We build AnalogNAS-Bench by extending NAS-Bench-201 and incorporating AIMC-specific constraints, allowing for a fair comparison of different architectures under \ac{HWT}. This provides a standardized framework for evaluating AIMC-aware NAS methodologies.
    \item We analyze the limitations of standard \ac{PTQ} \citep{banner2019post} and \ac{QAT} \citep{jacob2018quantization} in preserving network rankings under AIMC-induced noise, demonstrating that AIMC non-idealities introduce additional challenges that these methods do not address.
    \item Through systematic analysis on CIFAR-10 \citep{krizhevsky2009learning}, we extract key architectural insights that contribute to AIMC resilience. We study what makes certain architectures more robust to AIMC-induced errors, investigating factors such as network topology, connectivity patterns, and operator choices. Our findings provide guidelines for designing AIMC-optimized architectures and improving NAS methodologies for analog computing.
\end{enumerate}


The remainder of the paper is structured as follows: we first describe the AnalogNAS-Bench benchmark in Section~\ref{sec:analognas-bench-description}. We then analyze architectural robustness and extract key insights in Section~\ref{sec:analysis-insights}, followed by an initial comparison of NAS methodologies in Section~\ref{sec:initial-comparison}. Finally, we discuss limitations and future work in Section~\ref{sec:limitations}.

\section{AnalogNAS-Bench Description}
\label{sec:analognas-bench-description}

\begin{figure}[t]
  \centering
  \includegraphics[width=1\textwidth]{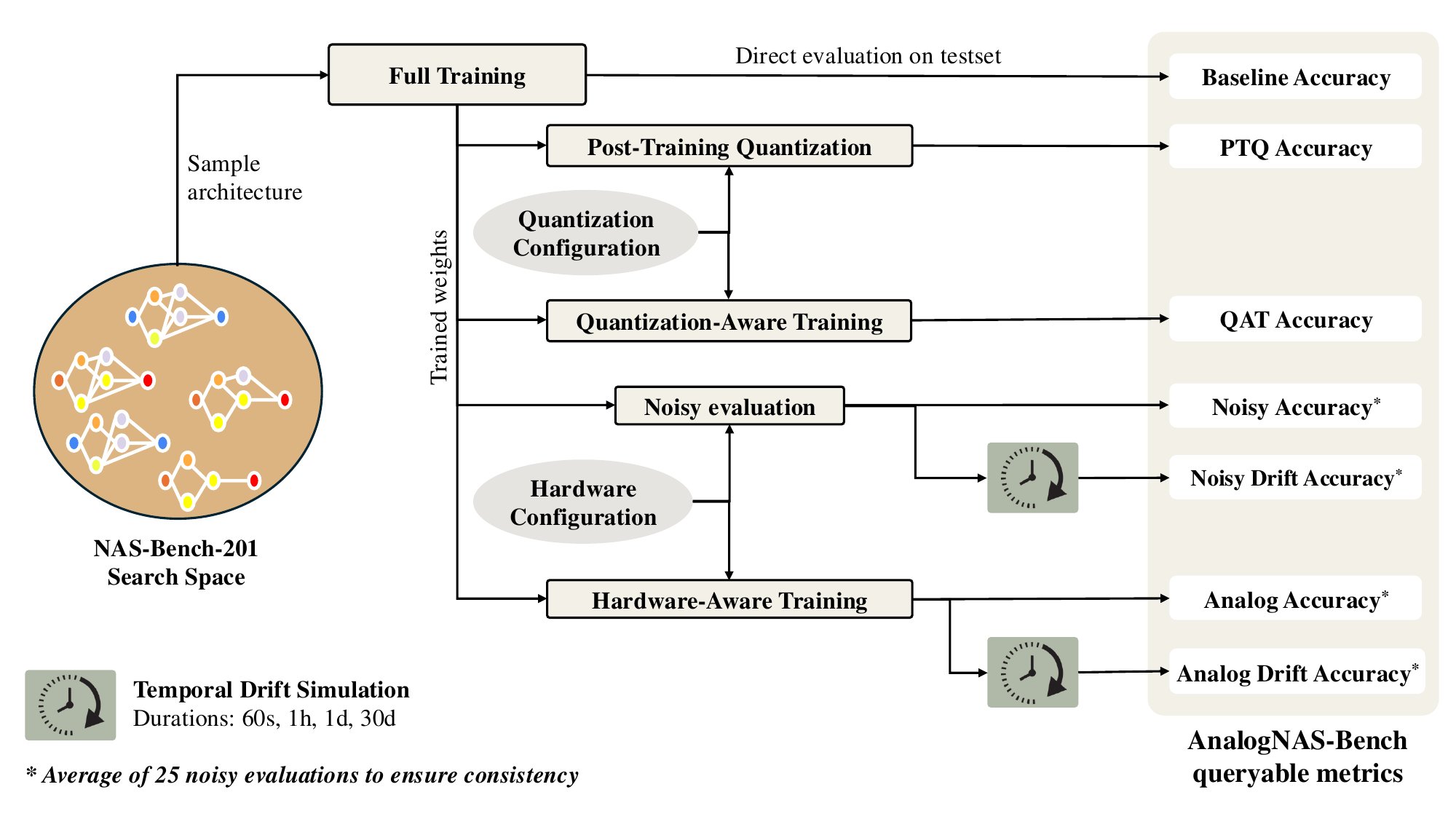}
  \caption{Overview of AnalogNAS-Benchmarking process.}
  \label{fig:overview}
\end{figure}
We construct our benchmark based on the search space of NAS-Bench-201 \citep{dong2020nasbench201}, a widely adopted \ac{NAS} benchmark. A full description can be found in Section \ref{sec:nb201} of the appendix. This search space is particularly well-suited for studying \ac{AIMC}-based inference due to its small but diverse range of architectures. Despite its architectural diversity, most models achieve high accuracy in standard digital hardware with low variance ($\text{median} = 90.41\%$, $\text{IQR} = 87.45\text{--}91.82\%$). This consistency allows for meaningful comparisons when evaluating the impact of analog constraints and supports the choice of analyzing this search space.

To systematically assess accuracy degradation and improvements under \ac{AIMC} settings, we define the following metrics, summarized in Figure \ref{fig:overview}:

\noindent
\textbf{Baseline Accuracy}— represents an architecture’s performance in a full-precision digital setting without hardware constraints. This metric serves as an upper bound.

\noindent
\textbf{Noisy Accuracy}— measures the accuracy degradation that occurs when an architecture is deployed on \ac{AIMC} hardware without \ac{HWT}. We compute this by mapping digital-trained weights onto an \ac{AIMC} crossbar and simulating inference using IBM’s AIHWKit \citep{aihwkit}, which incorporates noise sources from real memristive devices.

\noindent
\textbf{Analog Accuracy}— quantifies the performance of an architecture trained with hardware-aware optimization, where \ac{AIMC}-specific noise and non-idealities are incorporated into the training process. Prior research \citep{joshi2020accurate, rasch2023hardware, yang2022tolerating} has shown that integrating \ac{AIMC} constraints during training enables models to develop compensatory mechanisms, enhancing robustness against hardware-induced distortions. The Analog Accuracy metric thus serves as an indicator of the effectiveness of \ac{HWT} in mitigating \ac{AIMC}-induced errors and optimizing performance in analog computing environments.

In addition to static accuracy metrics, we evaluate model robustness under drift conditions. \textbf{Noisy Drift} represents the accuracy of a model deployed directly onto AIMC hardware and subjected to drift effects over different time intervals. The degradation is measured at \textbf{60 seconds},\textbf{ 1 hour}, \textbf{24 hours}, and \textbf{30 days}. Similarly, \textbf{Analog Drift} quantifies the accuracy degradation for models trained with \ac{HWT}. Users can easily use our API to query
the results of any architecture.

\section{Analysis \& Insights}\label{sec:analysis-insights}
In this section, we analyze the performance of neural network architectures on \acs{AIMC}, focusing on key hardware-induced constraints: \textbf{quantization}, \textbf{analog noise}, and \textbf{drift}. We examine how these factors affect accuracy, assess the role of \acs{HWT} in mitigating performance degradation, and identify architectural characteristics that contribute to robustness. We aim to define the structural features that enable architectures to perform reliably under \acs{AIMC} constraints. We present here the main insights, more analysis can be found in Section~\ref{sec:graf} of the appendix.

\subsection{Architecture Ranking}  
\begin{figure}[t]
  \begin{subfigure}[t]{0.5\linewidth}
    \centering
    \includegraphics[scale=0.47]{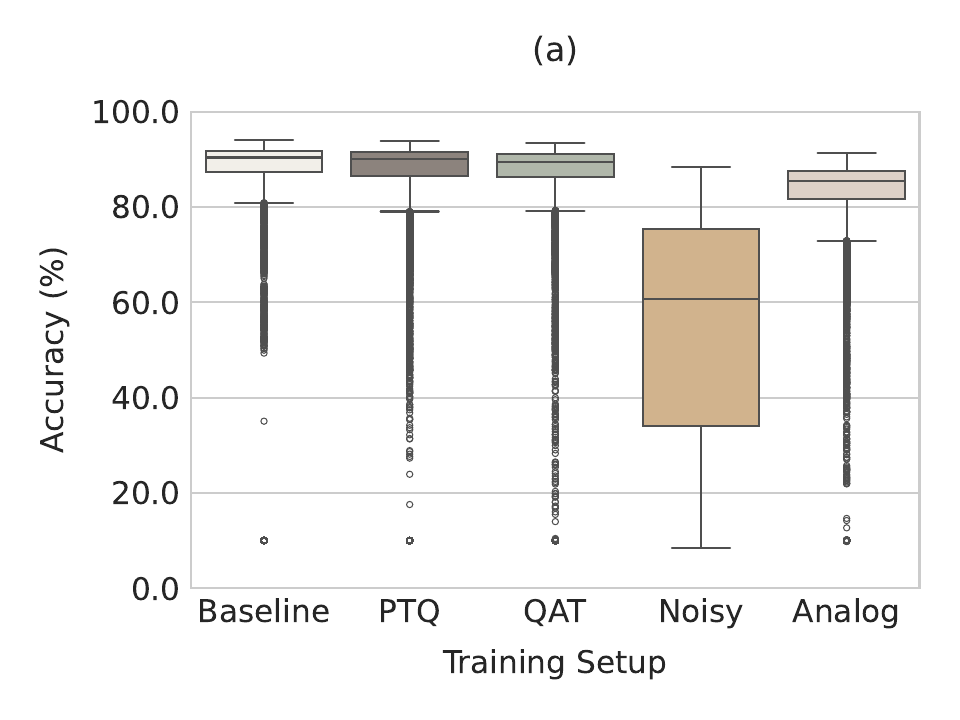}
  \end{subfigure}
  \begin{subfigure}[t]{0.5\linewidth}
    \centering
    \includegraphics[scale=0.42]{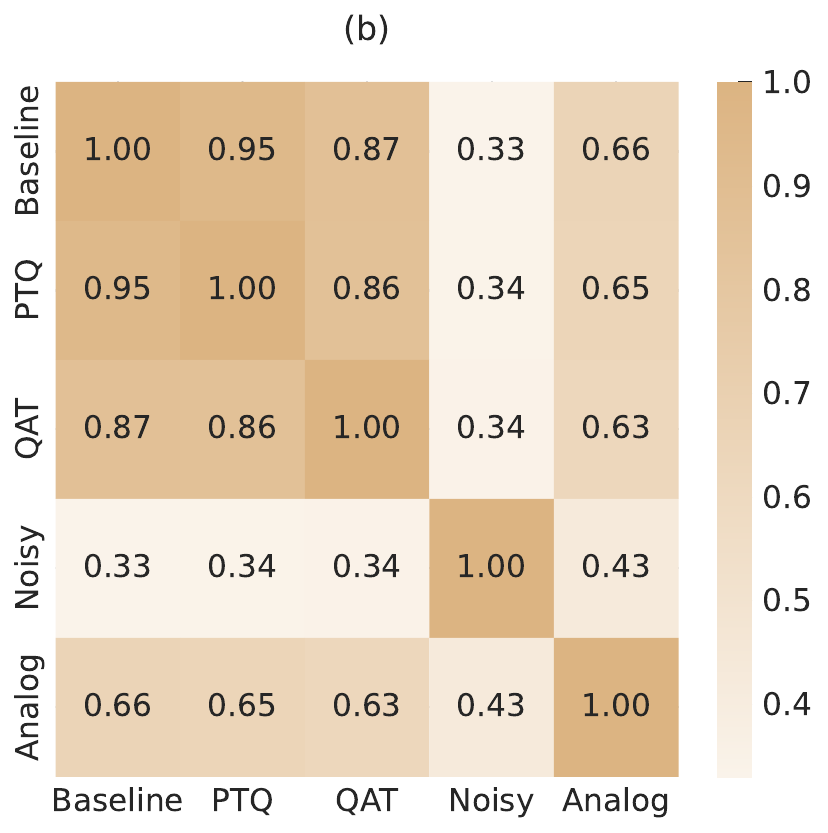}
  \end{subfigure}
  \caption{(a) Accuracy distribution and (b) Kendall’s tau correlation heatmap  across Baseline, PTQ, QAT, Noisy, and Analog training setups on CIFAR-10.}  
  \label{fig:acc-rank}
\end{figure}

To evaluate architectural performance across different training and deployment settings, we analyze accuracy rankings and correlations (Figure~\ref{fig:acc-rank}). \textbf{Noisy performance shows significant variability}, with a large accuracy spread ($\text{std} = 25.47\%$) and a lower median ($60.70\%$). The interquartile range ($33.99\text{--}75.39\%$) highlights substantial robustness differences across architectures. Weak correlation between noisy accuracy and baseline accuracies ($\tau = 0.33$) suggests that performance on standard digital hardware does not reliably predict \acs{AIMC} behavior. Similarly, quantization robustness does not imply \acs{AIMC} robustness, as \acs{PTQ} and \acs{QAT} also show weak correlation with noisy accuracy ($\tau = 0.34$). These results indicate that \textbf{analog noise is the primary factor driving performance degradation}. This decline, however, is mitigated as \acs{HWT} improves accuracy significantly, raising the mean to $81.31\%$ and the median to $85.49\%$. The stronger correlation between baseline and analog accuracies ($\tau = 0.66$) confirms that \acs{HWT} reduces noise-induced losses. However, analog accuracy remains below baseline levels ($\text{IQR} = 81.70\text{--}87.55\%$ vs. $87.45\text{--}91.82\%$), showing that \textbf{\acs{AIMC} noise imposes fundamental constraints that \acs{HWT} cannot fully overcome in certain architectures.}

\subsection{Noise Robustness Analysis}
To identify architectures that are naturally robust to analog noise, we analyze its direct impact on accuracy. Specifically, we focus on architectures that achieve high baseline accuracy ($>90\%$) and evaluate how their accuracy degrades under \acs{AIMC} noise. We define \textbf{noisy drop} as the difference between baseline and noisy accuracies. Architectures are then categorized as robust or non-robust based on a threshold set by the lower quartile ($12.75\%$) of the noisy drop.

\subsubsection*{Operation Distribution}
\begin{figure}[t]
  \centering
  \includegraphics[width=0.8\linewidth]{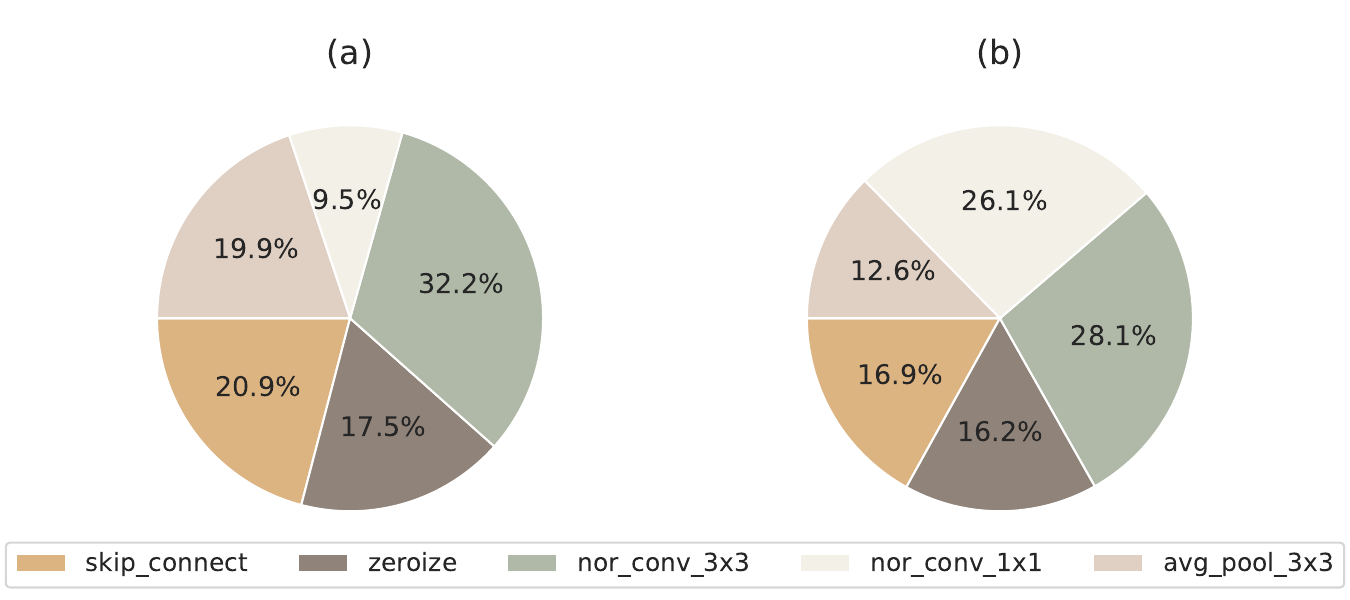}
  \caption{Percentage of each operation type in (a) robust and (b) non-robust architectures.}
  \label{fig:op-dist}
\end{figure}

Figure~\ref{fig:op-dist} shows the mean percentage of each operation type in robust and non-robust architectures. A clear distinction emerges: \textbf{robust} architectures contain \textbf{far fewer 1$\times$1 convolutions} ($9.50\%$ vs. $26.10\%$ in non-robust ones), as these operations utilize less of the crossbar array, making them more susceptible to additional noise. Instead, they \textbf{favor 3$\times$3 convolutions ($32.18\%$)}, which leverage larger crossbar regions. Additionally, \textbf{higher occurrences of average pooling} ($19.88\%$) \textbf{and skip connections} ($20.93\%$) suggest that pooling helps reduce noisy features, while skip connections duplicate the information of the actual input and emphasize its importance. 

\subsubsection*{Operation Counts}
\begin{figure}[t]
    \centering
    \includegraphics[width=\linewidth]{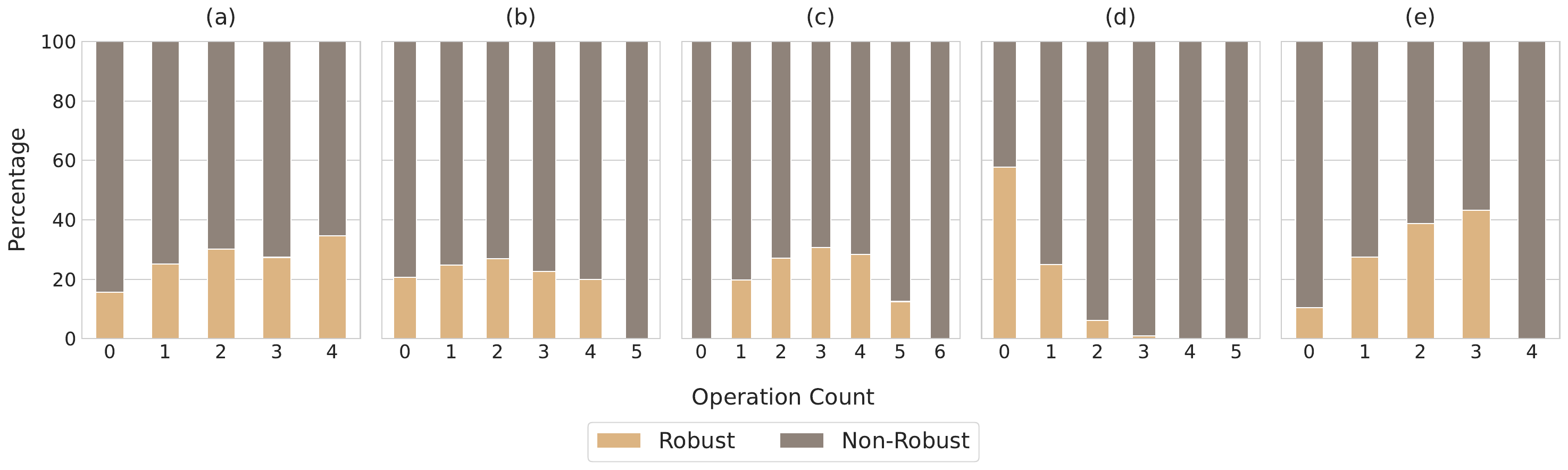}
    \caption{Distribution of operation counts for robust and non-robust architectures. From left to right: (a) skip connection, (b) zeroize, (c) 3$\times$3 convolution, (d) 1$\times$1 convolution, and (e) 3$\times$3 average pooling.}
    \label{fig:op-count}
\end{figure}

Beyond these general trends, robustness is strongly linked to the specific counts of each operation. Figure~\ref{fig:op-count} presents the proportion of architecture groups for varying occurrences of each operation. When architectures \textbf{lack 1$\times$1 convolutions}, more than half are robust, but robustness quickly declines as their count increases. In contrast, \textbf{adding multiple 3$\times$3 convolutions} consistently improves robustness, though the effect plateaus beyond four. \textbf{Pooling and skip connections also exhibit an optimal range}: architectures with two or three of either operation tend to be more robust, while having too few or too many reduces performance. Meanwhile, \textbf{the zeroize operation shows no clear impact}, suggesting sparsity alone does not determine robustness.

\subsubsection*{Sequential Patterns}
\begin{figure}[t]
  \centering
  \includegraphics[scale=0.17]{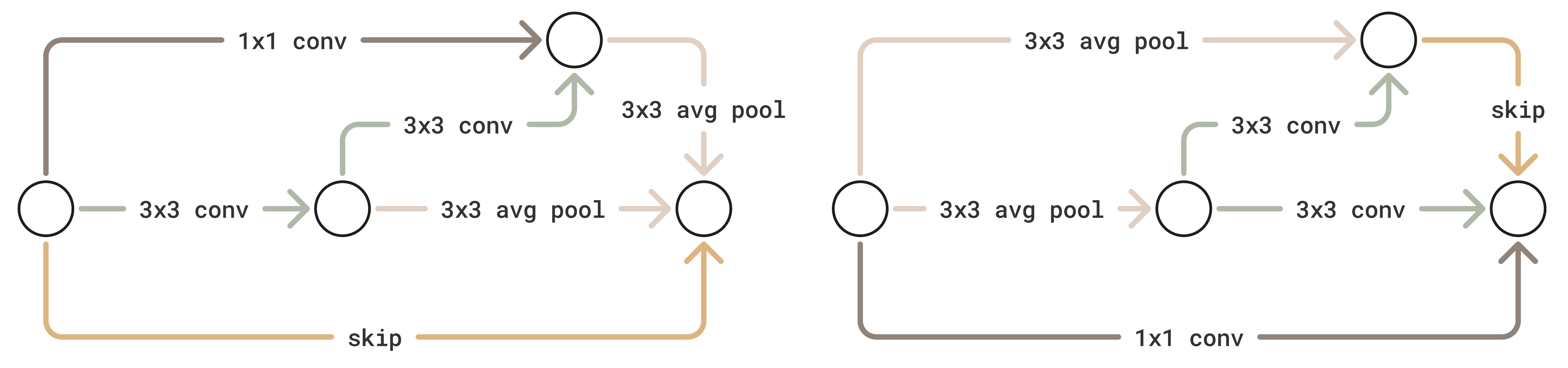}
  \caption{Two architectures with identical operation counts but different robustness. The left architecture (2, 3, 0, 2, 4, 4) achieves high noisy accuracy ($86.02\%$), while the right one (4, 4, 3, 2, 2, 0) performs poorly (only $44.27\%$).}
  \label{fig:ex-path-diff}
\end{figure}

Operation counts alone do not fully explain robustness, as architectures with identical compositions can exhibit vastly different robustness levels. This suggests that the order and positioning of operations, beyond just their frequencies, play a crucial role. An example of this effect is shown in Figure~\ref{fig:ex-path-diff}. To better understand these sequential patterns, we examine the most frequent one-, two-, and three-operation paths in robust and non-robust architectures. \textbf{Robust pathways consistently feature 3$\times$3 convolutions combined with skip connections or pooling layers}, as seen in sequences like (2,0), (2,4), (2,2,4), and (2,0,2). In contrast, \textbf{non-robust pathways frequently include 1$\times$1 convolutions}, appearing in patterns like (2,3), (3,2,3), and (2,3,3), indicating that their presence propagates instability \textbf{even when paired with other beneficial operations}.

\begin{figure}[t]
  \centering
  \includegraphics[scale=0.45]{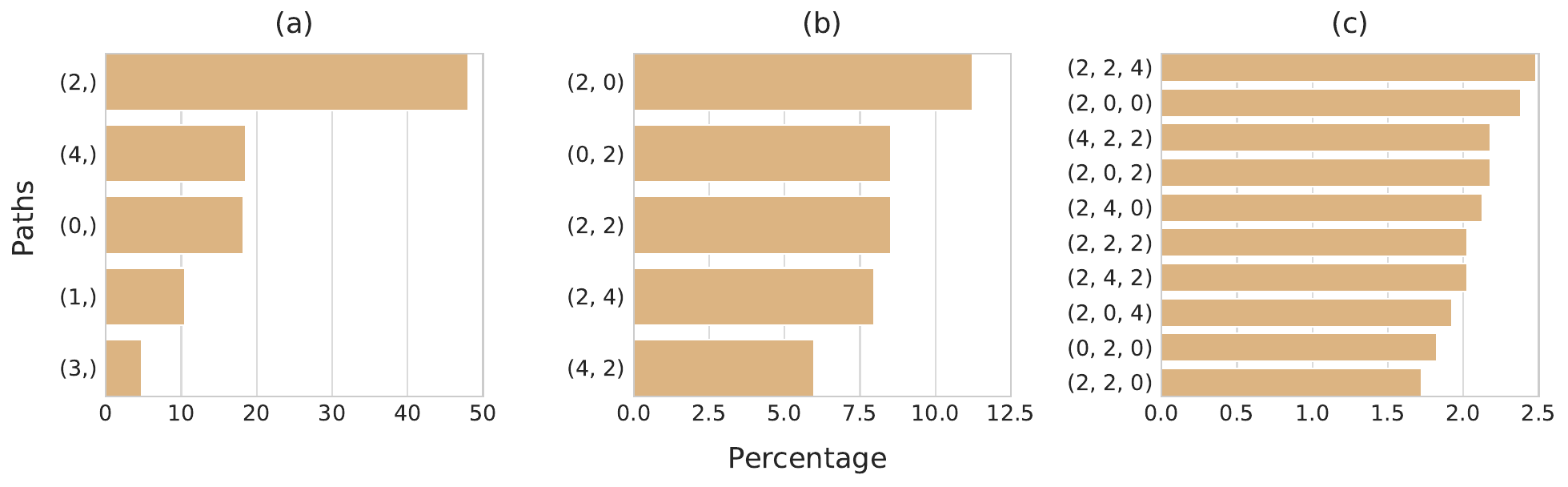}
  \caption{Most common paths of length (a) one, (b) two and (c) three in robust architectures. Operations are labeled as follows: (0) skip connection, (1) zeroize, (2) 3$\times$3 convolution, (3) 1$\times$1 convolution, and (4) 3$\times$3 average pooling.}
  \label{fig:op-path}
\end{figure}

\subsection{Hardware-Aware Training Robustness}
To evaluate the impact of \acs{HWT} on robustness, we analyzed the performance shift from noisy inference to analog execution. The majority of architectures achieved strong performance post-\acs{HWT}, with $80.19\%$ achieving an analog accuracy above $80\%$. Most architectures benefited significantly, with an average improvement of $134.17\%$, and $73.81\%$ of them showed substantial gains. However, some architectures remained highly susceptible to analog noise despite HWT interventions.

\subsubsection*{HWT vs. Natural Robustness}
\begin{figure}[t]
    \centering
    \includegraphics[width=0.7\textwidth]{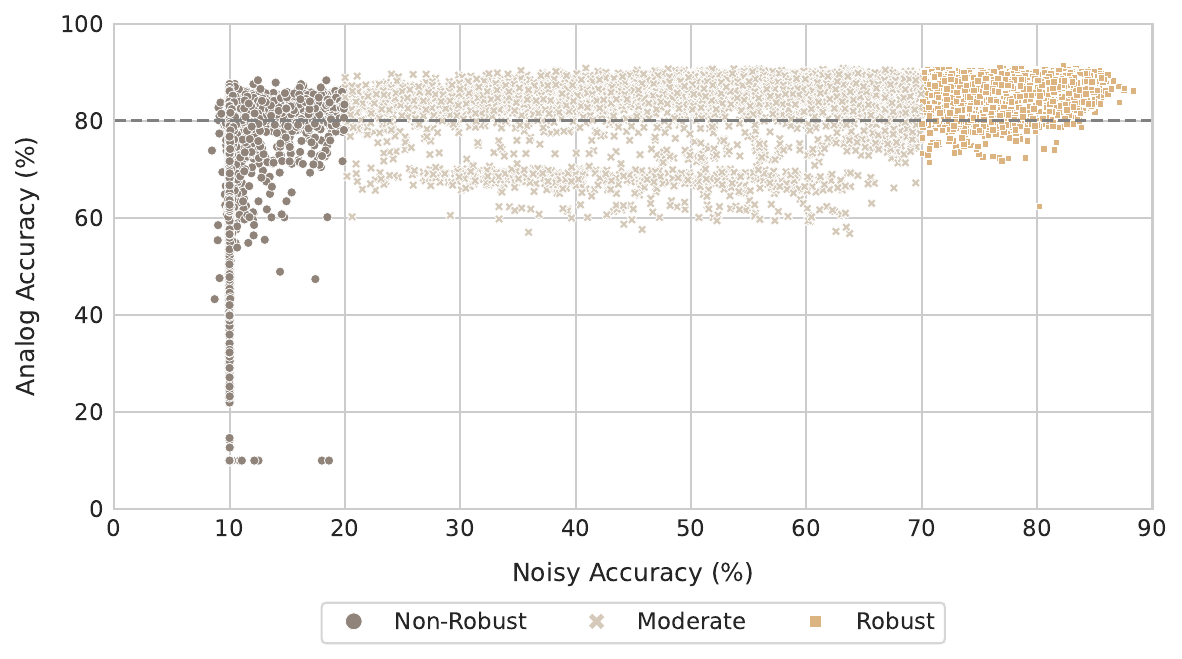}
    \caption{Impact of \acs{HWT} on accuracy across different architecture categories. Architectures are grouped as Non-Robust, Moderate, or Naturally Robust. The dashed line at 80\% marks the high analog accuracy threshold.}
    \label{fig:vs}
\end{figure}

To assess whether architectures that naturally resist analog noise perform better than those that significantly improve through \acs{HWT}, we categorized architectures into three groups based on their \textbf{noisy accuracy}: \textbf{Naturally robust} ($>70\%$), \textbf{Non-Robust} ($<20\%$), and \textbf{Moderate} ($20\%\text{--}70\%$). We then evaluated their analog accuracy (post-\acs{HWT}), setting a threshold of $80\%$ to define high-performing architectures (Figure~\ref{fig:vs}). Naturally robust architectures maintained strong performance, with $82.88\%$ exceeding $85\%$ analog accuracy and requiring minimal improvement ($\sim 10\%$) due to their inherent resilience. In contrast, non-robust architectures exhibited the highest variance ($\text{mean} = 65.72\%, \text{std} = 23.53\%$), with some failing entirely. However, $34.41\%$ of this group saw dramatic gains ($\text{mean improvement} = 495.99\%$), though most remained within $80\text{--}85\%$ range, , and only $4.63\%$ surpassed the most frequent analog accuracy observed in the naturally robust group. The moderate group showed mixed results: $86.67\%$ crossed the $80\%$ threshold, while some remained below $70\%$. Ultimately, \textbf{naturally robust architectures are the most reliable}, as they consistently perform well with minimal intervention, whereas \textbf{\acs{HWT} does not always recover all noise-sensitive architectures.}

\subsubsection*{HWT-Failing Architectures}
We examined the subset of the most underperforming architectures: those that initially performed well on standard hardware ($\text{baseline accuracy} >90\%$) but degraded severely under noise ($\text{noisy accuracy} < 20\%$) and failed to recover post-\acs{HWT} ($\text{analog accuracy} <40\%$). These architectures shared two key traits. First, they exhibited \textbf{excessive reliance on 1$\times$1 convolutions} ($31.25\%$) while \textbf{underutilizing 3$\times$3 convolutions} ($13.33\%$), particularly in early-stage connections, resulting in weak noise suppression due to insufficient filtering. Second, they \textbf{relied heavily on non-learnable operations}, which in many cases formed entire paths on their own ($33.75\%$), particularly in short pathways dominated by skip connections ($40\%$) and average pooling ($37.5\%$). Longer sequences also combined these non-learnable operations with 1$\times$1 convolutions ($37.75\%$), offering little resistance to noise propagation.

\subsection{Temporal Drift}
To assess the impact of temporal drift on neural architectures, we first establish accuracy degradation thresholds for different time durations. Specifically, we analyze performance drops at 60 seconds, 1 hour, 1 day, and 30 days under two distinct conditions: \textbf{Noisy Drift} (i.e., without \ac{HWT}) and \textbf{Analog Drift} (i.e., with \ac{HWT}). These thresholds are determined statistically by computing key distribution metrics, including the mean, 25th percentile, and 75th percentile of the accuracy drop of only architectures that perform well in digital (Baseline Accuracy > 80) and are noise resilient (Noisy Accuracy > 70). Architectures that maintain accuracy within these thresholds are classified as \textbf{robust}, while those exhibiting significant degradation are labeled \textbf{non-robust}. Full statistics and thresholds can be found in Section \ref{sec:temp_drift_stats} of the appendix.

\subsubsection*{Noisy Drift over time}
Robust architectures show a consistent \textbf{dominance of 3x3 convolutions}, which become even more prominent as the duration increases. \textbf{The skip connection operation also remains significant, indicating that robust architectures favor a balance between feature extraction and identity mapping}. In contrast, 1x1 convolutions appear minimally in robust architectures, suggesting that fine-grained transformations are less useful for maintaining robustness. However, \textbf{in non-robust architectures, 1x1 convolutions and zeroize operations tend to be more present}. 
Interestingly, as time progresses, robust architectures seem to increase their reliance on 3x3 convolutions while decreasing their use of 1x1 convolutions. This aligns with the hypothesis that \textbf{wider receptive fields contribute to robustness}, as 3x3 convolutions can better capture spatial correlations, making the model less susceptible to noise.

\subsubsection*{Analog Drift over time}
\begin{figure}[t]
    \centering
    \includegraphics[scale=0.44]{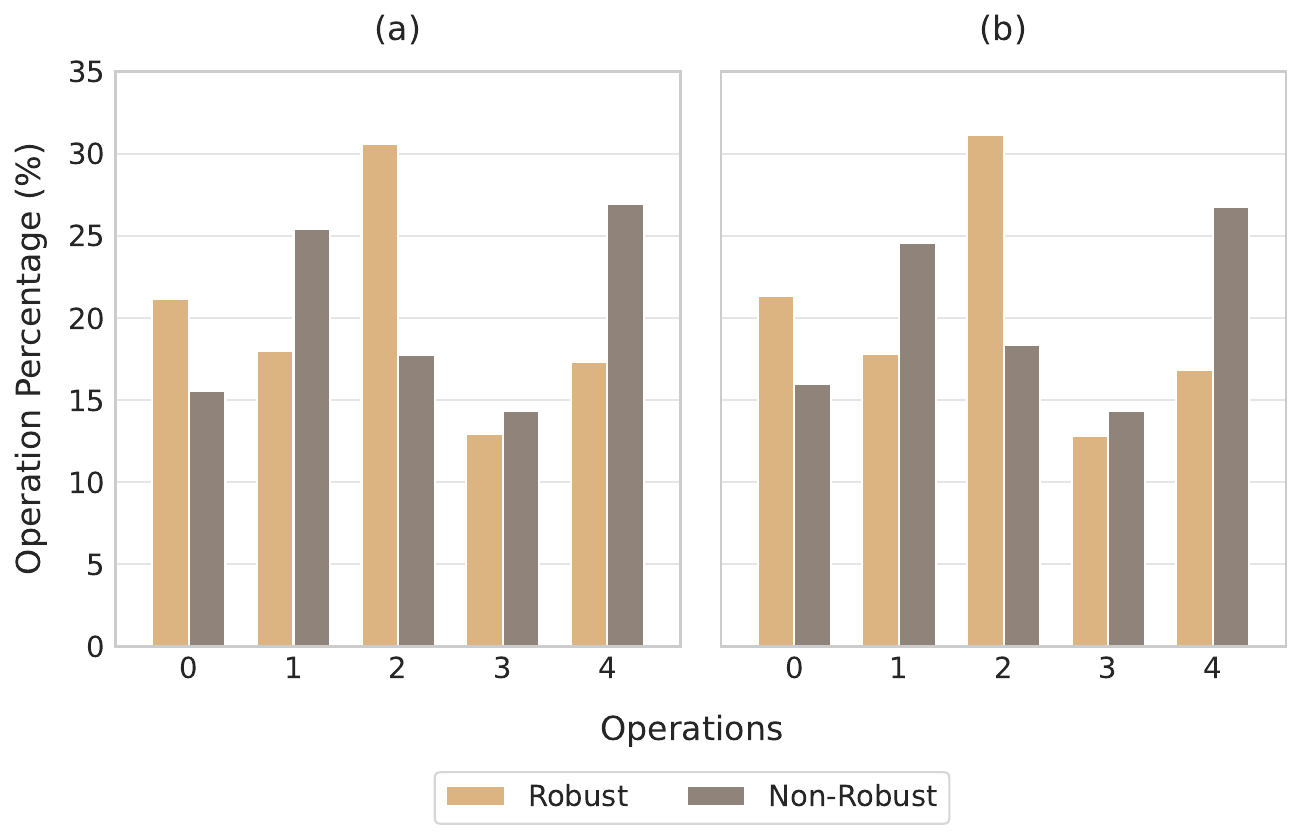}
    \caption{Operation presence over time in (a) Robust and (b) Non-Robust architectures trained with \ac{HWT}. Operations are: (0) skip connection, (1) zeroize, (2) 3$\times$3 convolution, (3) 1$\times$1 convolution, and (4) 3$\times$3 average pooling.}
    \label{fig:vs-drift}
\end{figure}

Figure \ref{fig:vs-drift} demonstrates that \textbf{robust architectures show a higher presence of skip connections and 3×3 convolutions, suggesting that these operations contribute to long-term stability.} The dominance of 3×3 convolutions indicates that spatial feature extraction plays a crucial role in resilience to analog drift. In contrast, \textbf{1×1 convolutions remain minimal}. \textbf{On the other hand, Non-robust architectures exhibit a higher presence of average pooling and zeroize operations—both of which lack trainable parameters.} This reliance on non-learnable operations likely makes them more vulnerable to analog drift, as \ac{HWT} offers no direct mechanism to counterbalance the absence of weight adaptability. While both robust and non-robust architectures maintain similar patterns, the gap in operation presence increases, suggesting that \textbf{robustness characteristics become more pronounced over longer periods}.

\section{Initial Comparison}\label{sec:initial-comparison}
In this section, we present an initial benchmark evaluating standard NAS methodologies using AnalogNAS-Bench. Table~\ref{tab:nas} summarizes the performance of the best architectures identified by various search strategies, primarily focusing on maximizing the 1-day accuracy. This comparison establishes a baseline for NAS methods tailored specifically to AIMC-based hardware. Except for BANANAS \citep{white2021bananas}, NAS4RRAM \citep{yuan2021nas4rram}, NACIM \citep{jiang2021nacim}, and Gibbon \citep{sun2023gibbon}—which employ their own specialized surrogate modeling techniques—all other methods utilize an XGBoost \citep{chen2016xgboost} surrogate model trained on a dataset comprising 900 architectures from the AnalogNAS-Bench search space. The surrogate approach aims to capture realistic search times in practical scenarios.

The results in Table~\ref{tab:nas} demonstrate that among AIMC-specific methods, AnalogNAS \citep{benmeziane2023analognas} and GA for IMC AI hardware \citep{ga} achieve the highest 1-day accuracy (90.04\%), nearly matching the performance of the optimal architecture obtained via exhaustive search (90.53\%). General NAS methods, including Random Search and Evolutionary Algorithms, also achieve competitive baseline accuracies; however, they tend to exhibit higher variability in accuracy over one month (AVM). Notably, Bayesian Optimization and BANANAS show lower accuracies and higher AVM, highlighting their limited effectiveness when analog objectives are considered. AIMC-specific approaches consistently demonstrate improved stability and higher robustness against noise, reinforcing the importance of analog-aware search strategies. These findings highlight a promising direction for future research: developing more suitable estimation strategies that simultaneously consider stability (AVM) and robustness (1-day accuracy), thereby potentially enhancing the effectiveness of NAS methods tailored specifically for AIMC hardware.

\begin{table}[!t]
\centering
\caption{Benchmark Comparison of NAS Methodologies on AnalogNAS-Bench}
\label{tab:nas}
\begin{threeparttable}
\resizebox{\textwidth}{!}{%
\begin{tabular}{lcccccc}
\toprule\toprule
\textbf{Method} & \textbf{Baseline Acc.} & \textbf{Noisy Acc.} & \textbf{1-day Acc.} & \textbf{AVM*} & \textbf{\# Parameters (K)} & \textbf{Search Time (s)} \\
\midrule
\multicolumn{7}{c}{\textbf{Best Architecture}} \\
\midrule
Exhaustive Search & 93.46 & 82.27 & 90.53 ± 0.28 & 0.79 ± 0.03 & 802 & - \\
\midrule
\multicolumn{7}{c}{\textbf{General NAS Methods}} \\
\midrule
Random Search & 93.34 & 85.10 & 89.93 ± 0.25 & 3.71 ± 0.14 & 802  & 201 \\
Evolutionary Algorithm & 93.32 & 82.21 & 89.34 ± 0.12 & 0.59 ± 0.11 & 834 & 621 \\
Bayesian Optimization & 90.21 & 80.10 & 82.65 ± 1.54 & 6.43 ± 1.07 & 766  & 764 \\
BANANAS \citep{white2021bananas} & 91.31 & 80.24 & 85.97 ± 1.12 & 5.91 ± 0.87 & 803 & 872 \\
\midrule
\multicolumn{7}{c}{\textbf{AIMC-Specific Methods}} \\
\midrule
AnalogNAS \citep{benmeziane2023analognas} & 93.12 & 84.37 & 90.04 ± 0.31 & 2.13 ± 0.22 & 567 & 530 \\
NAS4RRAM \citep{yuan2021nas4rram} & 92.45 & 83.51 & 89.12 ± 0.42 & 1.97 ± 0.19 & 856 & 612 \\
GA for IMC AI hardware \citep{ga} &  93.12 & 84.37 & 90.04 ± 0.31 & 2.13 ± 0.22 & 567  & 689 \\
NACIM\textsuperscript{†} \citep{jiang2021nacim} & 92.65 & 82.95 & 89.30 ± 0.45 & 1.11 ± 0.20 & 877 & 725 \\
Gibbon\textsuperscript{†} \citep{sun2023gibbon} & 92.79 & 83.04 & 89.45 ± 0.40 & 1.34 ± 0.18 & 936 & 698 \\
\bottomrule\bottomrule
\end{tabular}}
\begin{center}
\begin{tablenotes}
\footnotesize
\item[*] AVM: Accuracy Variation over one Month
\item[†] NACIM and Gibbon jointly optimize network and hardware parameters. \\ Hardware optimization is omitted from this comparison.
\end{tablenotes}
\end{center}
\end{threeparttable}
\end{table}

\section{Limitations \& Future work}\label{sec:limitations}

While the AnalogNAS-Bench provides a comprehensive platform to analyze convolutional neural networks under AIMC constraints, several limitations must be highlighted. First, the current benchmark evaluations are fully completed only on the CIFAR-10 dataset. Although experiments with CIFAR-100 \citep{krizhevsky2009learning} and ImageNet16-120 \citep{imagenet16} are actively underway, the generalization of insights and robustness trends across more complex and diverse datasets remains an essential next step. Moreover, the current benchmark contains a relatively small set of architectures (derived from NAS-Bench-201). While beneficial for systematic analysis, this limited search space also creates an opportunity: the dataset size is suitable for developing zero-cost estimation methods to quickly predict architecture robustness against analog noise, facilitating exploration of substantially larger and more diverse search spaces in the future.

\noindent
Second, our benchmark is limited to convolution-based network architectures. Although these architectures naturally align with analog crossbar implementations, in their im2col \citep{chellapilla2006high} format, transformer-based models have recently gained significant attention due to their extensive matrix-vector multiplications and large parameter counts, which can benefit substantially from AIMC stationary computations, i.e., avoiding weight loading \citep{spoon2021toward, sridharan2023x}. Thus, incorporating a dedicated transformer-based search space into the benchmark is crucial for extracting insights into architectural robustness specific to transformer models.

\noindent Lastly, all current evaluations assume a fixed hardware configuration, detailed in the appendix, and are primarily based on Phase-Change Memory (PCM) device characteristics for noise analysis. Future extensions should explore mappings of neural architectures onto diverse analog hardware platforms, including Resistive RAM (RRAM), Ferroelectric RAM (FeRAM), Electrochemical RAM (ECRAM), and other emerging devices \citep{joshi2020accurate}. Additionally, expanding to heterogeneous systems that combine analog and digital hardware components will provide deeper insights into the architecture-hardware interplay. Exploring these variations will significantly broaden the applicability and robustness insights of AnalogNAS-Bench across multiple analog computing scenarios.

\section{Conclusion}

In this paper, we introduced AnalogNAS-Bench, the first NAS benchmark tailored for \ac{AIMC}, extending NAS-Bench-201 with AIMC-specific constraints to enable a fair evaluation of architectures under analog noise conditions, bridging a critical gap in existing \ac{NAS} benchmarks that overlook \ac{AIMC} non-idealities. Through our analysis, we highlighted the limitations of conventional quantization techniques in capturing AIMC-specific noise and identified key architectural traits that enhance resilience: increased skip connections, wider layers, and a preference for larger convolutional operations over narrower ones. With this benchmark in place, more NAS algorithms and estimators will be developed by researchers, making it more practical to compare them in AIMC scenarios.

%
%
%
%
%

\begin{acknowledgements}
We thank the computational support from AiMOS, an AI supercomputer made available by the IBM Research AI Hardware Center and Rensselaer Polytechnic Institute’s Center for Computational Innovations (CCI).
\end{acknowledgements}


\bibliography{references}




\newpage

\newpage
\appendix


\section{Related works}


\paragraph{Hardware-aware Neural Architecture Search (HW-NAS) for \ac{AIMC}}
HW-NAS aims to find the most efficient DNN for a specific dataset and target hardware platform. Many works \citep{benmeziane2023analognas, yuan2021nas4rram, ga, jiang2021nacim, sun2023gibbon} target \ac{AIMC} using HW-NAS. \textbf{AnalogNAS} is a NAS framework designed specifically for \ac{AIMC}. It incorporates hardware constraints, noise modeling and an evolutionary search strategy. \textbf{NAS4RRAM} is a method to find an efficient \ac{DNN} for a specific RRAM-based accelerator. It uses an evolutionary algorithm, trains each sampled architecture without \ac{HWT} training, and evaluates each network on a specific hardware instance. \textbf{NACIM} uses co-exploration strategies to find the most efficient architecture and the associated hardware platform. For each sampled architecture, networks are trained considering noise variations. This approach is limited by using a small search space due to the high time complexity of training. \textbf{Gibbon} is a co-exploration framework for neural network models and \ac{AIMC} architectures, which utilizes an evolutionary search algorithm with adaptive parameter priority to optimize performance and energy efficiency.


\section{NAS-Bench-201 Search Space}
\label{sec:nb201}
\begin{figure}[b]
  \centering
  \includegraphics[width=1\textwidth]{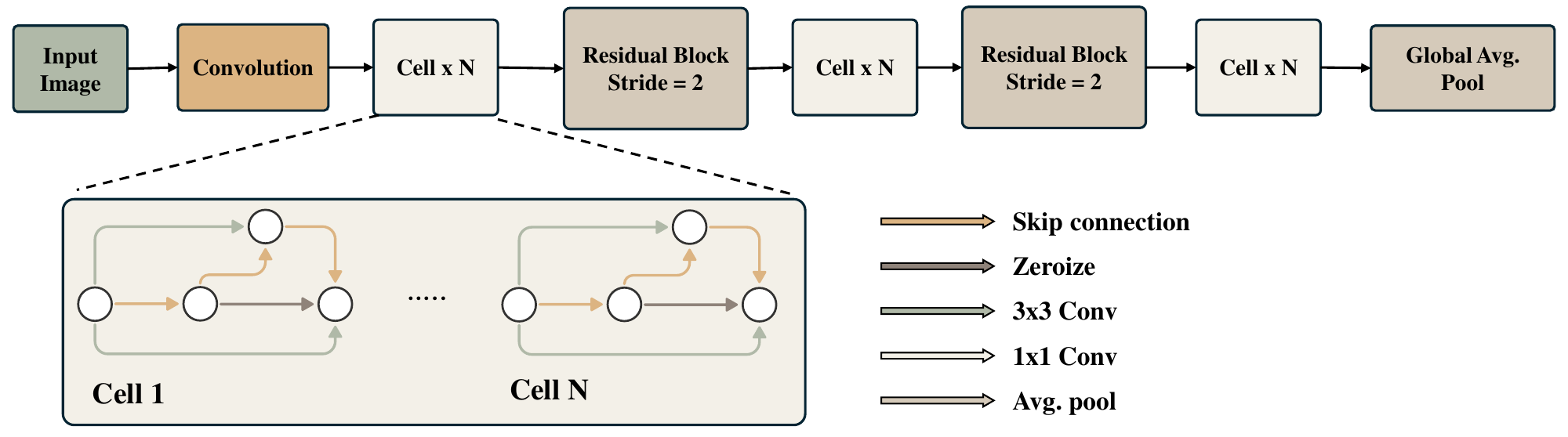}
  \caption{NAS-Bench-201 macro-achitecture.}
  \label{fig:nb201}
\end{figure}

NAS-Bench-201 defines a compact yet diverse set of architectures, each represented as a \ac{DAG}. The search space consists of networks with a fixed macro-structure, where the primary design choices lie within a single cell repeated multiple times throughout the architecture. Each cell consists of a 4-node structure, where every edge represents one of five possible operations: (0) Skip connection (or Identity), (1) zeroize, (2) 3$\times$3 convolution, (3) 1$\times$1 convolution, or (4) 3$\times$3 average pooling (Figure~\ref{fig:nb201}). An architecture can be represented and encoded using the cell. The search space contains 15,625 architectures, each of which has been trained and evaluated on standard datasets, including CIFAR-10, CIFAR-100, and ImageNet16-120.

\section{Experimental Setup}
To evaluate neural architectures under \ac{AIMC} constraints, we conduct large-scale experiments using a high-performance computing cluster equipped with NVIDIA Tesla V100 GPUs. 
Architectures are sampled from the NAS-Bench-201 search space, and we utilize NASLib \citep{naslib-2020} to extract them as PyTorch models. 

The training pipeline is implemented in PyTorch, with SLURM managing job scheduling across multiple GPUs. Given the large number of architectures—15,625 in total—we employ a distributed training strategy where each architecture is mapped to a dedicated GPU in a batched manner. All architectures are trained for 200 epochs using \ac{SGD} with Nesterov momentum, set to 0.9. A cosine annealing learning rate schedule \citep{loshchilov2017sgdr} is applied, decaying the initial learning rate of 0.1 to 0 over the course of training. To prevent overfitting, weight decay is set to $5 \times 10^{-4}$. In addition, standard data augmentation techniques are employed to improve generalization. The training pipeline includes random horizontal flipping with a probability of 0.5, random cropping to 32 × 32 pixels with 4-pixel padding, and RGB normalization to standardize input features. These preprocessing steps ensure consistency across different architectures and datasets.

Model evaluation is performed under multiple conditions to assess both digital and analog performance. Different accuracies (Baseline, \ac{PTQ}, \ac{QAT}, Noisy, Analog, Drift) are computed as the mean over 25 independent evaluation runs on the test set. \ac{PTQ} models are obtained by quantizing weights and activations to int8 without additional fine-tuning, whereas \ac{QAT} training is conducted using Adam optimizer with a learning rate of 0.001 and a weight decay of $1 \times 10^{-4}$, with the ReduceLROnPlateau scheduler applied to dynamically adjust the learning rate based on performance improvements.

To simulate analog hardware, we use IBM’s AIHWKit framework. Different noise levels are defined through custom hardware configurations, allowing precise control over hardware non-idealities. Table \ref{tab:rpu_config} provides a detailed description of the hardware configurations used in our experiments.

\begin{table}[t]
  \caption{Hardware Configuration used for experiments in aihwkit}
  \label{tab:rpu_config}
  \centering
  \begin{tabular}{ll}
    \toprule
    \textbf{Parameter} & \textbf{Setting} \\
    \midrule
    Digital-to-Analog Converter (DAC) & 8 bits \\
    Analog-to-Digital Converter (ADC) & 8 bits \\
    Output noise & $\sigma = 0.04$ \\
    Max. conductance & 25 $\mu$S \\
    Programming noise & $\sigma = 1$ \\
    Read noise & $\sigma = 1$ \\
    Global drift compensation & Enabled \\
    \bottomrule
  \end{tabular}
\end{table}

Experiments are conducted on three benchmark datasets of increasing complexity. Training and evaluation on CIFAR-10 have been completed, while experiments on CIFAR-100 and ImageNet16-120 are ongoing. These datasets allow us to evaluate architecture generalization across different classification tasks and ensure that our findings extend beyond a single dataset.

\section{Quantization Performance}
\acs{QAT} and \acs{PTQ} exhibit a strong correlation ($\tau = 0.86$), indicating similar effects on architectures. Their high correlation with baseline accuracy ($\tau = 0.95$ for QAT, $\tau = 0.87$ for PTQ) suggests that \textbf{standard training transfers well to quantized models with minimal ranking disruption}. Moreover, the tight distribution overlap further supports that quantization is a manageable constraint for most architectures. Notably, \textbf{\acs{PTQ} alone appears sufficient for performance estimation}, offering an efficient alternative to \acs{QAT}.

\section{Temporal Drift Statistics}
Table~\ref{tab:drift_stats} summarizes the statistics and thresholds of the different drift metrics.
In the case of noisy drift, we aim to identify architectures that remain robust without \ac{HWT}, which means they are naturally robust. To set a stricter criterion, we define the robustness threshold between the minimum drop and the first quartile, ensuring that only the most resilient architectures are selected.
On the other hand, For analog drift, our goal is to identify architectures that remain non-robust even with \ac{HWT}. Since most architectures recover with adaptation, we set the robustness threshold between the third quartile and the maximum drop, capturing those that still degrade significantly despite adaptation.
\label{sec:temp_drift_stats}
\begin{table}[t]
    \caption{Statistical summary of drift metrics}
    \label{tab:drift_stats}
    \centering
    \begin{tabular}{lcccccccc}
        \toprule
        \textbf{Metric (\%)}  & \textbf{Mean} & \textbf{Std} & \textbf{Min} & \textbf{25\%} & \textbf{50\%} & \textbf{75\%} & \textbf{Max} & \textbf{Threshold} \\
        \midrule
        Noisy Drift Drop 60s  & 14.33 & 6.72 & 1.90 & 9.54 & 13.05 & 17.76 & 50.75 & \textbf{5} \\
        Noisy Drift Drop 1h  & 24.63 & 7.96 & 6.67 & 18.84 & 23.66 & 29.59 & 69.02 & \textbf{10} \\
        Noisy Drift Drop 1d  & 35.76 & 8.66 & 12.97 & 29.63 & 35.52 & 41.85 & 70.20 & \textbf{16} \\
        Noisy Drift Drop 30d  & 47.54 & 7.87 & 21.26 & 42.18 & 48.26 & 53.36 & 70.92 & \textbf{25} \\
        \midrule
        Analog Drift Drop 60s  & 1.59 & 1.39 & -1.34 & 0.70 & 1.20 & 2.07 & 17.61 & \textbf{2.5} \\
        Analog Drift Drop 1h  & 2.32 & 2.02 & -0.61 & 1.05 & 1.74 & 2.94 & 32.01 & \textbf{3.5} \\
        Analog Drift Drop 1d  & 3.44 & 2.96 & -0.46 & 1.59 & 2.53 & 4.29 & 36.95 & \textbf{4.5} \\
        Analog Drift Drop 30d  & 5.27 & 4.40 & -0.13 & 2.49 & 3.86 & 6.49 & 44.89 & \textbf{7} \\
        \bottomrule
    \end{tabular}
\end{table}

\section{Neural Graph Features Analysis}
\label{sec:graf}
To better understand their structure, neural network architectures can be analyzed using graph-based features in terms of node connectivity, path relationships, and the presence of specific operations \citep{graf}. These features can be grouped into several categories: \textbf{operation count (op\_count)}, which measures the frequency of specific operations; \textbf{minimum path length (min\_path\_len)}, which captures the shortest distance between nodes; \textbf{maximum operations on a path (max\_op\_on\_path)}, which identifies the most frequent operations along a path; and \textbf{node degree metrics (node\_degree)}, which include \textbf{in-degree}, \textbf{out-degree}, and \textbf{average degree} measures that describe how connected specific operations are within the architecture. Each of these features is computed for different subsets of operations within the search space. In this section, we analyze how these features correlate with robustness and provide insights into their impact on architectural design.

\subsection{Noisy Robustness}
\begin{figure}[t]
  \centering
  \includegraphics[width=1\textwidth]{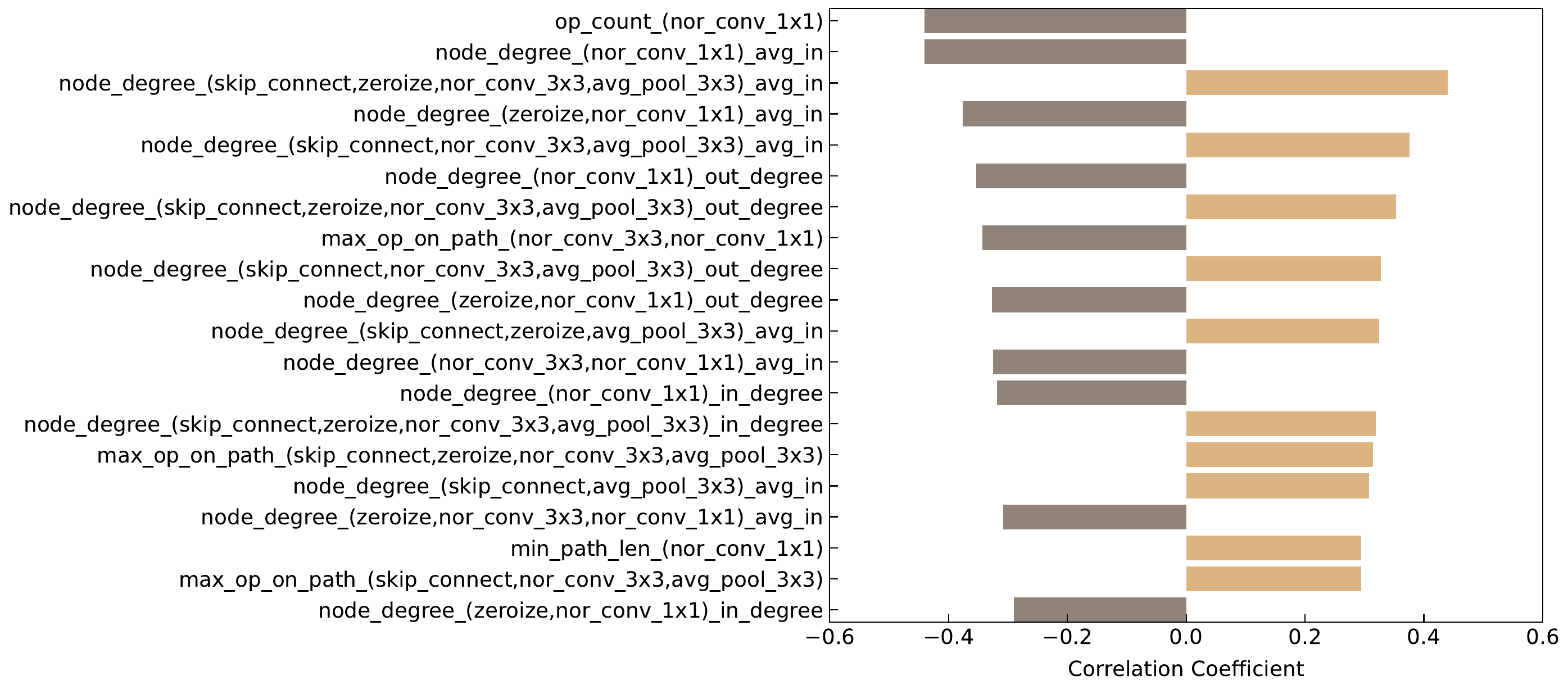}
  \caption{Top 20 features correlated with noisy robustness.}
  \label{fig:graf-noisy}
\end{figure}

Figure \ref{fig:graf-noisy} highlights the correlation of graph features with noisy robustness. Higher node degree across diverse operations is associated with increased robustness, while path-based features suggest that well-connected, structured pathways contribute to stability. In terms of operations, architectures that heavily rely on $1\times1$ convolutions tend to be less robust, whereas those incorporating skip connections, $3\times3$ convolutions, and average pooling in key positions demonstrate greater resilience.

\subsection{HWT Robustness}
\begin{figure}[t]
  \centering
  \includegraphics[width=1\textwidth]{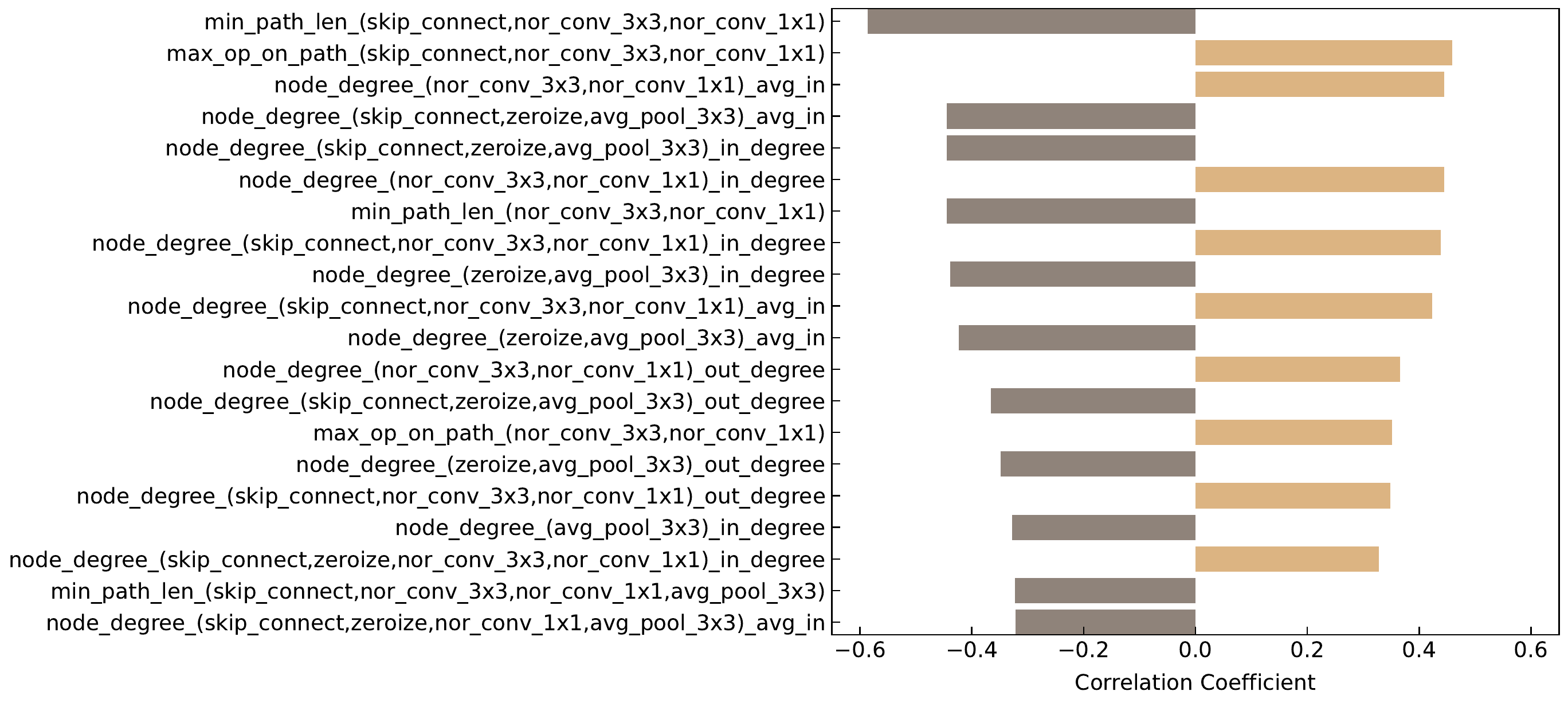}
  \caption{Top 20 features correlated with HWT robustness.}
  \label{fig:graf-hwt}
\end{figure}

Figure \ref{fig:graf-hwt} illustrates the correlation of graph features with HWT robustness impact. Longer paths and higher node degree, \textbf{when driven by convolutions}, suggesting that deep, well-connected architectures adapt better to transformation. However, pathways dominated by non-learnable operations such as pooling and zeroize show reduced improvement, indicating that static components limit adaptability. These results highlight that HWT is most effective in architectures with strong connectivity and learnable operations, while non-learnable elements hinder its impact.

\newpage
\subsection{Drift Robustness}
\begin{figure}[]
  \centering
  \includegraphics[width=\textwidth]{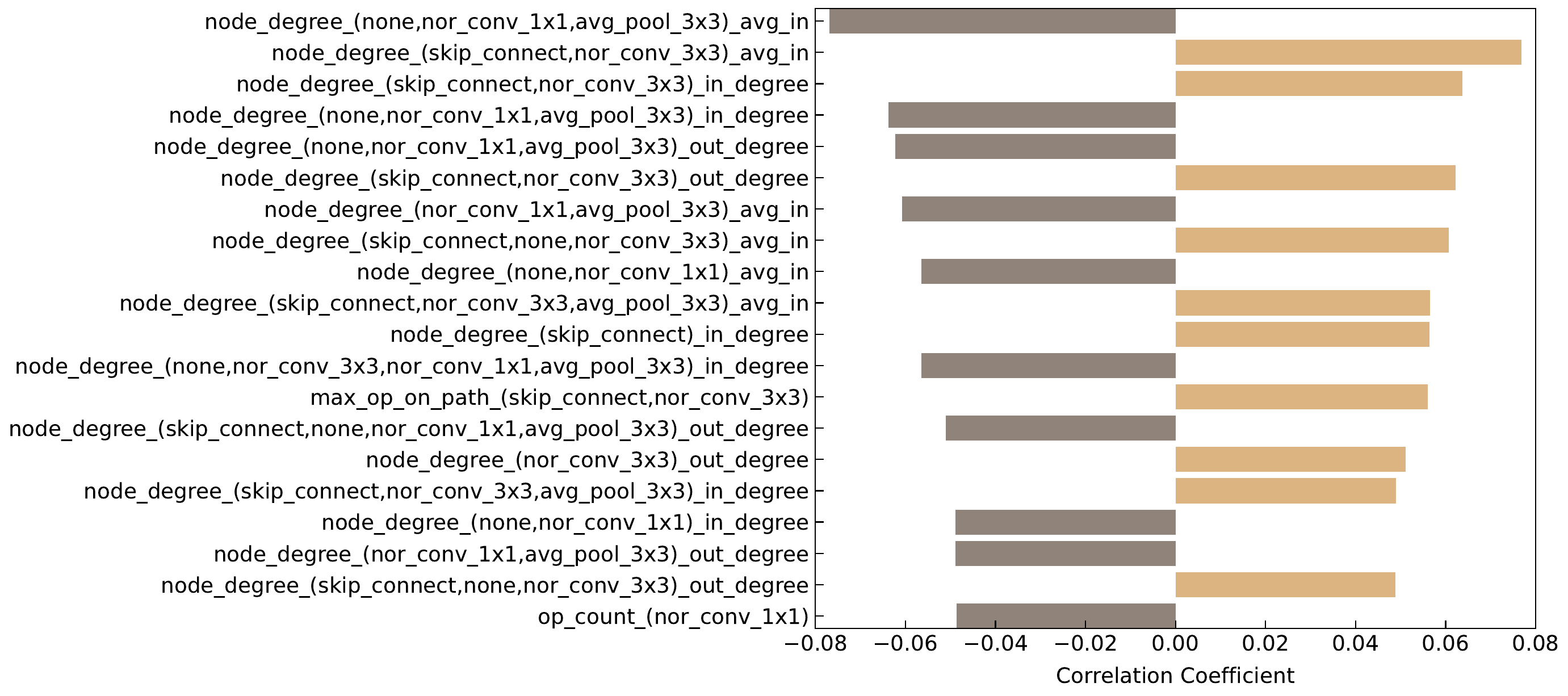}
  \caption{Top 20 features correlated with robustness over 1 Day for Noisy Drift (without \ac{HWT}).}
  \label{fig:graf-noisy-drift-1d}
\end{figure}

\begin{figure}[]
  \centering
  \includegraphics[width=\textwidth]{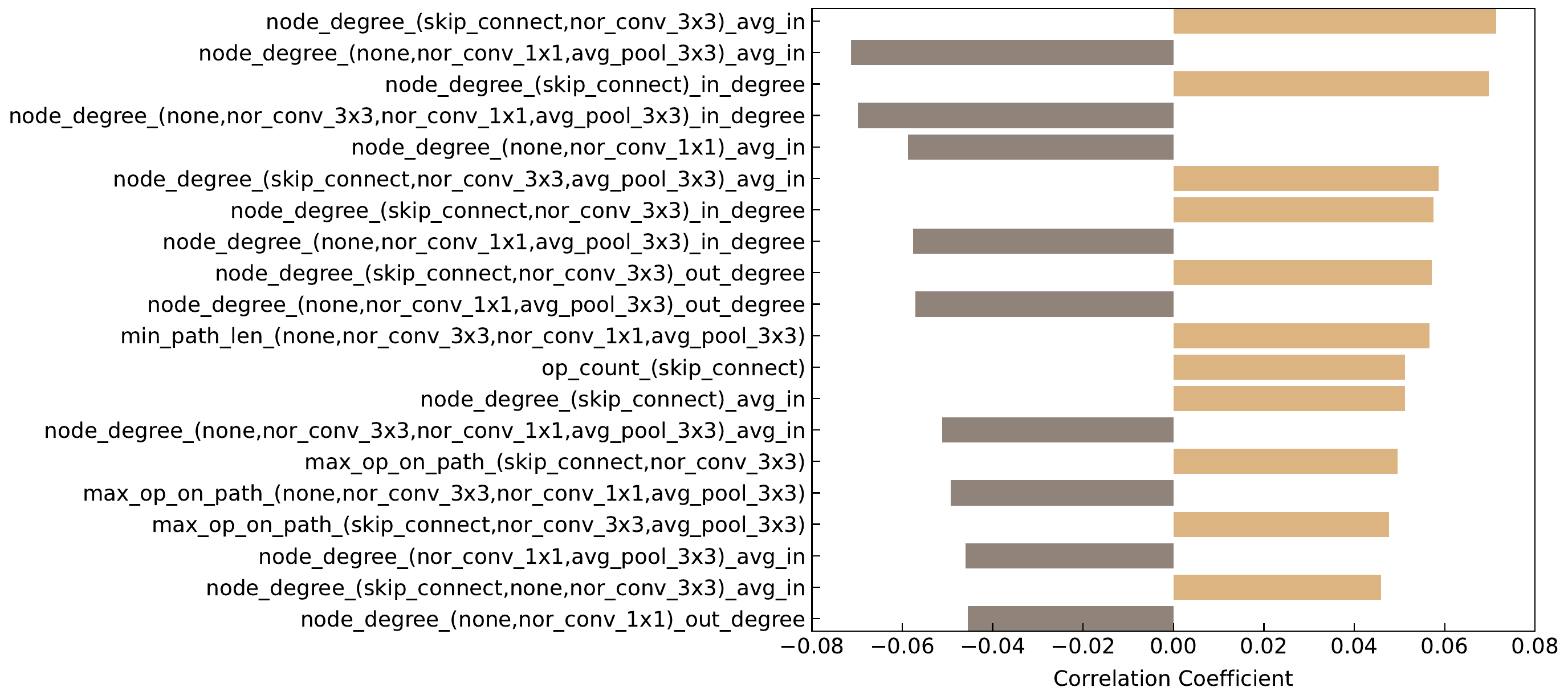}
  \caption{Top 20 features correlated with robustness over 1 Month for Noisy Drift (without \ac{HWT}).}
  \label{fig:graf-noisy-drift-1m}
\end{figure}

\begin{figure}[]
  \centering
  \includegraphics[width=\textwidth]{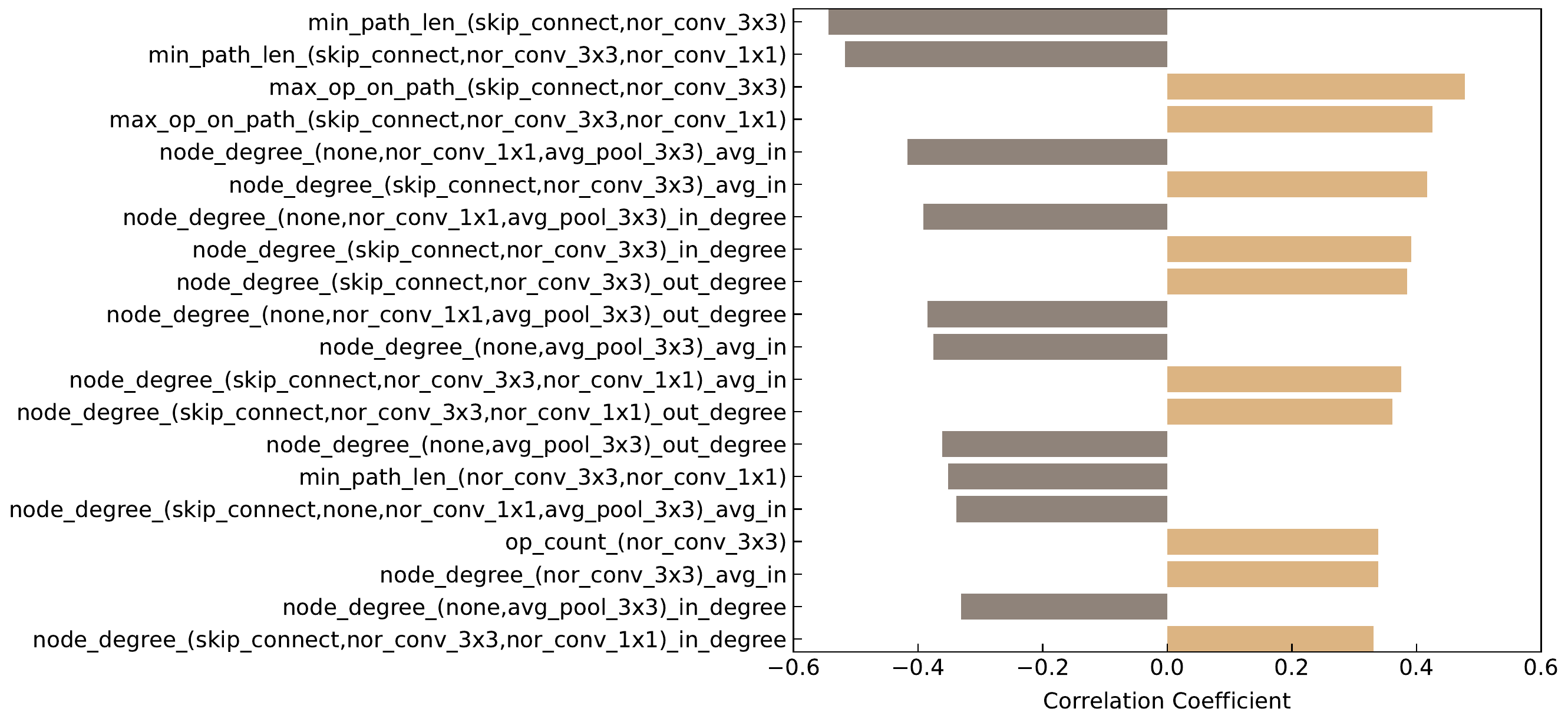}
  \caption{Top 20 features correlated with robustness over 1 Day for Analog Drift (with \ac{HWT}).}
  \label{fig:graf-analog-drift-1d}
\end{figure}

\begin{figure}[]
  \centering
  \includegraphics[width=\textwidth]{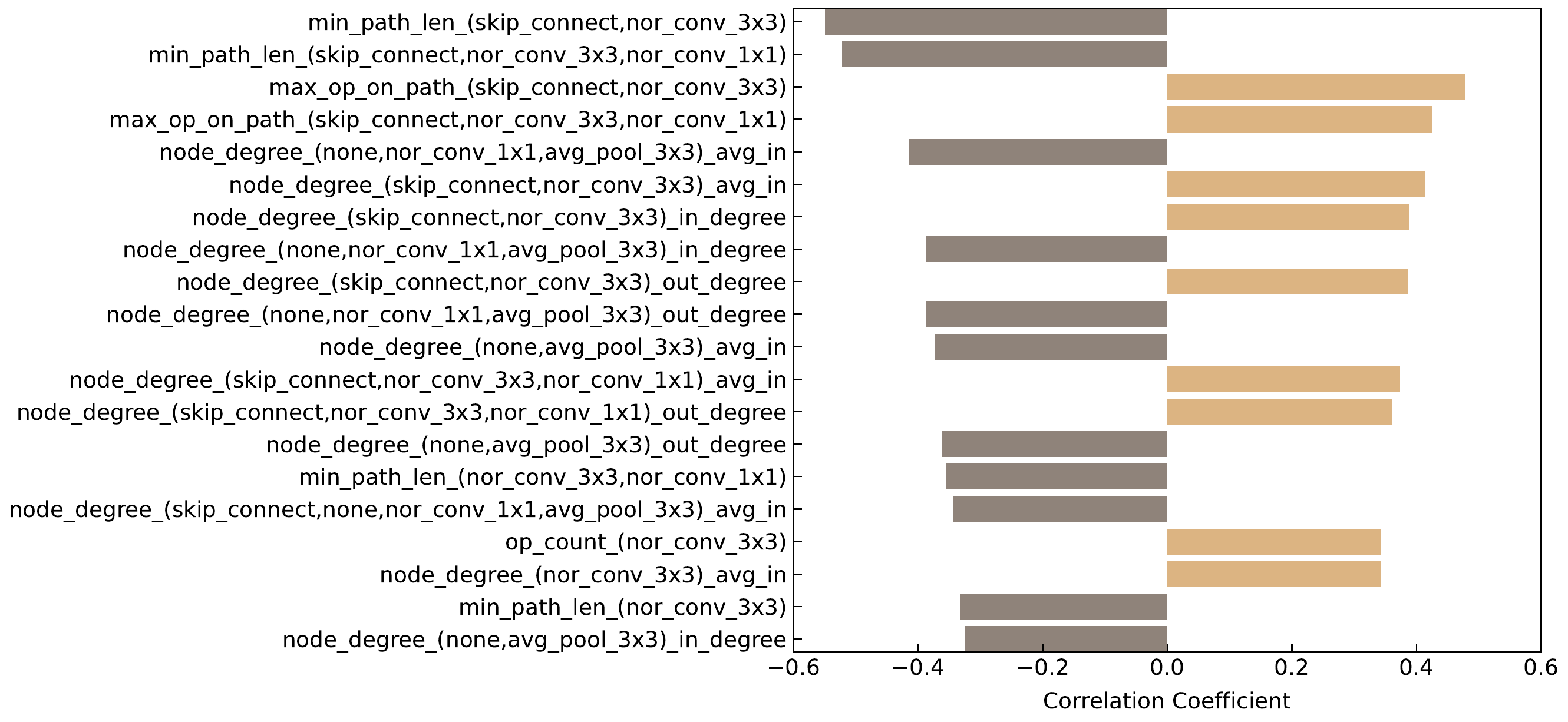}
  \caption{Top 20 features correlated with robustness over 1 Month for Analog Drift (with \ac{HWT}).}
  \label{fig:graf-analog-drift-1m}
\end{figure}

Figure~\ref{fig:graf-noisy-drift-1d} to \ref{fig:graf-analog-drift-1m} show the correlation of graph features with drift robustness.
In the case of Noisy Drift, the most influential feature in robustness prediction is node degree, which appears most frequently among the top 20 correlated features. Architectures with \textbf{a high node degree for} \textbf{skip connections and $3\times3$ convolutions} tend to demonstrate higher robustness. In contrast, architectures where the \textbf{node degree is high for $1\times1$ convolutions, none operations, and average pooling} are more likely to be non-robust.

For Analog Drift, the most significant features are related to path properties within the network. Robust architectures exhibit a \textbf{high maximum path length involving $3\times3$ convolutions and skip connections}, whereas non-robust architectures tend to have a \textbf{low minimum path length} for the same operations. 

\end{document}